\documentclass[
twocolumn,
]{ceurart}

\usepackage{listings}
\usepackage[utf8]{inputenc}
\usepackage{subfig}
\usepackage{multirow}
\begin{document}

\copyrightyear{2022}
\copyrightclause{Released under a custom license.}

%%


\conference{To be published at the Second Workshop on Multimodal Fact-Checking and Hate Speech Detection (DEFACTIFY'23) at the AAAI 2023 Conference,
February 14, 2023, Washington, D.C.}

\title{Fine-grained Czech News Article Dataset: \\ An Interdisciplinary Approach to Trustworthiness Analysis}

%\tnotetext[1]{You can use this document as the template for preparing your
%  publication. We recommend using the latest version of the ceurart style.}

%
% The "author" command and its associated commands are used to define
% the authors and their affiliations.
\author[1,2]{Matyáš Boháček}[%
orcid=0000-0001-8683-3692,
email=matyas.bohacek@matsworld.io,
url=www.matyasbohacek.com,
]

\address[1]{Faculty of Social Sciences, Charles University, Prague, Czech Republic}
\address[2]{Gymnasium of Johannes Kepler, Prague, Czech Republic}

\author[1,3]{Michal Bravanský}[%
orcid=0000-0002-2603-3017,
email=michal@bravansky.com,
url=www.bravansky.com/research,
]

\address[3]{University College London, United Kingdom}

\author[1,3]{Filip Trhlík}[%
orcid=0000-0003-3118-2911,
email=me@trhlikfilip.com,
url=www.trhlikfilip.com/research,
]

\author[1]{Václav Moravec}[%
orcid=0000-0002-3349-0785,
email=vaclav.moravec@fsv.cuni.cz,
url=iksz.fsv.cuni.cz/en/contacts/people/86190856,
]

% Footnotes

\begin{abstract}
  We present the Verifee Dataset: a novel dataset of news articles with fine-grained trustworthiness annotations. We develop a detailed methodology that assesses the texts based on their parameters encompassing editorial transparency, journalist conventions, and objective reporting while penalizing manipulative techniques. We bring aboard a diverse set of researchers from social, media, and computer sciences to overcome barriers and limited framing of this interdisciplinary problem. We collect over $10,000$ unique articles from almost $60$ Czech online news sources. These are categorized into one of the $4$ classes across the credibility spectrum we propose, raging from entirely trustworthy articles all the way to the manipulative ones. We produce detailed statistics and study trends emerging throughout the set. Lastly, we fine-tune multiple popular sequence-to-sequence language models using our dataset on the trustworthiness classification task and report the best testing F-1 score of $0.52$. We open-source the dataset, annotation methodology, and annotators' instructions in full length at \url{https://verifee.ai/research} to enable easy build-up work. We believe similar methods can help prevent disinformation and educate in the realm of media literacy.
\end{abstract}


\maketitle

\section{Introduction}

Donald Trump has called journalists and news outlets “fake news” nearly $2,000$ times since the beginning of his presidency, averaging more than one daily broadside against the press between 2016 and 2020~\cite{woodward2020fake}. Because of Trump, the term fake news underwent a fundamental change in its meaning. At first, it referred to a satirical and ironic genre of fictional news designed to entertain the audience. The original “fake news” have appeared on TV shows such as Saturday Night Live on NBC or in print, such as The Onion. However, during Trump’s campaign for the US presidential election in 2016 and his presidency, the concept of fake news became an integral part of his political communication. It aimed to discredit critical journalistic content or the whole news media as “fake media.” The successful stigmatization strategy of “fake news” has led to a fascination with this phenomenon in the public discourse and science. 

Fake news has become a label for false news and a synonym for both disinformation and misinformation. This has strengthened the binary perception of the credibility of information in a true-false dichotomous perspective. However, this reductionist approach has become a barrier to understanding the more profound meaning that the buzzword “fake news” covers. If we want to examine the credibility of the news content seriously, it is not possible to adopt the binary approach of either truth or lie. By creating the Verifee Dataset, we try to overcome the interdisciplinary barrier between social sciences (especially journalism and media studies) and computer science. This barrier prevents specialists in automated or robotic journalism from adopting a more analytical approach to various types of information disorders that we have become used to labelling with the general term “fake news”.

\section{Related Work}

Herein, we first review the current literature focusing on disinformation and misinformation in the journalistic ambit. We later provide an overview of existing methods treating these phenomena within the Artificial intelligence (AI) and Natural language processing (NLP) research communities. We first list some of the already-available datasets and then focus on the architectures solving the tasks of fake news detection and automatic fact-checking. 

\begin{figure}[h!]
\begin{center}
\includegraphics[width=0.8\linewidth]{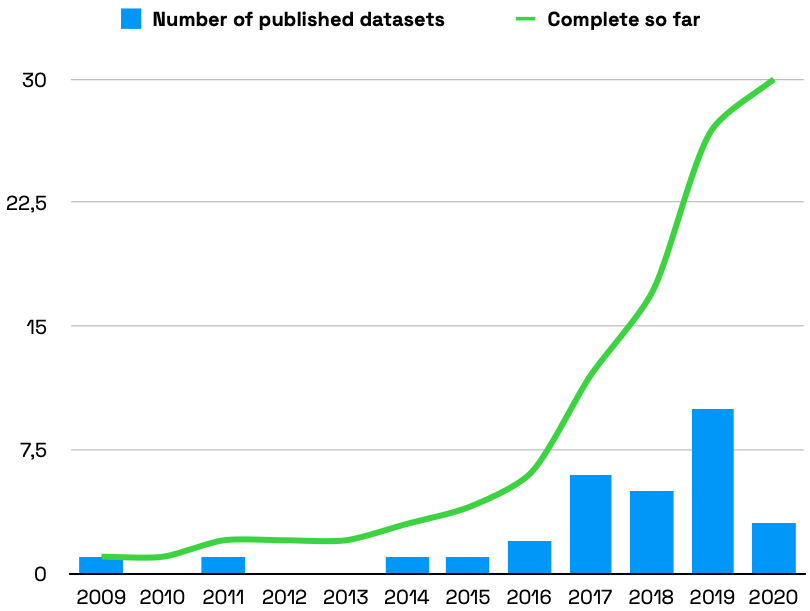}
\end{center}
\caption{Continual statistics on disinformation classification datasets publishing throughout the years 2009-2020. The bar charts denote the number of new datasets published in the respective year, while the overlay line captures the cumulative number of datasets published until that year. Years are plotted on the x-axis, and the numbers of published datasets are represented on the y-axis.}
\label{fig:dataset_growth}
\end{figure}

The task of fake news detection resides in classifying whether a given news article (or occasionally another medium such as a Tweet) is considered fake (disinformative) or truthful (credible). There is no consensus in the literature on what specific parameters derive from these states, but truthfulness is usually considered the primary one. Some approaches recognize more fine-grained scales with specific classes, such as tabloid news, mixed reliability news, whereas others only recognize fake and credible news. Either way, this task extracts the class from the text on its own.

The task of automatic fact-checking, on the other hand, requires a source of truth to which the news article is compared. The task then lies in determining whether the article is supported by facts therein. Hence, one can consider this task a specific abbreviation of stance detection focusing on news media and large-scale ground-truth databases.

We review datasets and approaches in both of these tasks, as our dataset lies somewhere in between.

\begin{figure*}[h]
  \centering
  \hspace{0.8cm}
  \subfloat[Number of classres recognized by the dataset]{\includegraphics[width=0.4\textwidth]{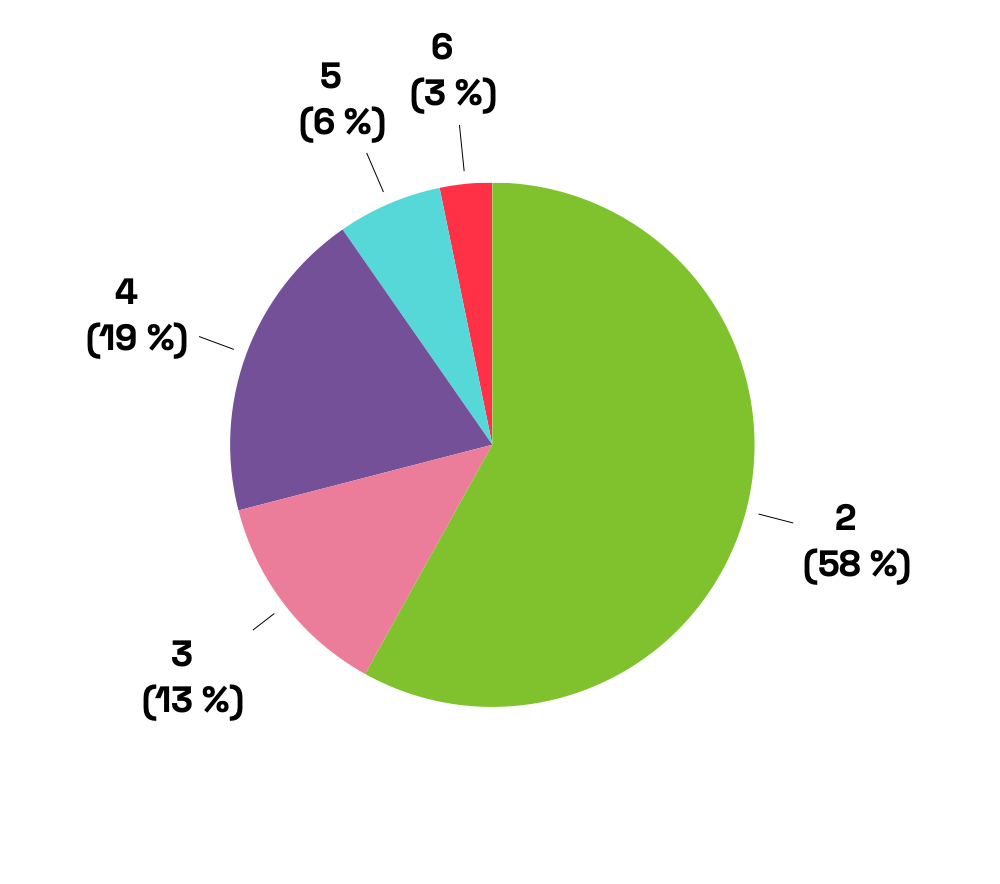}\label{fig:classes}}
  \hfill
  \subfloat[Language of the dataset]{\includegraphics[width=0.4\textwidth]{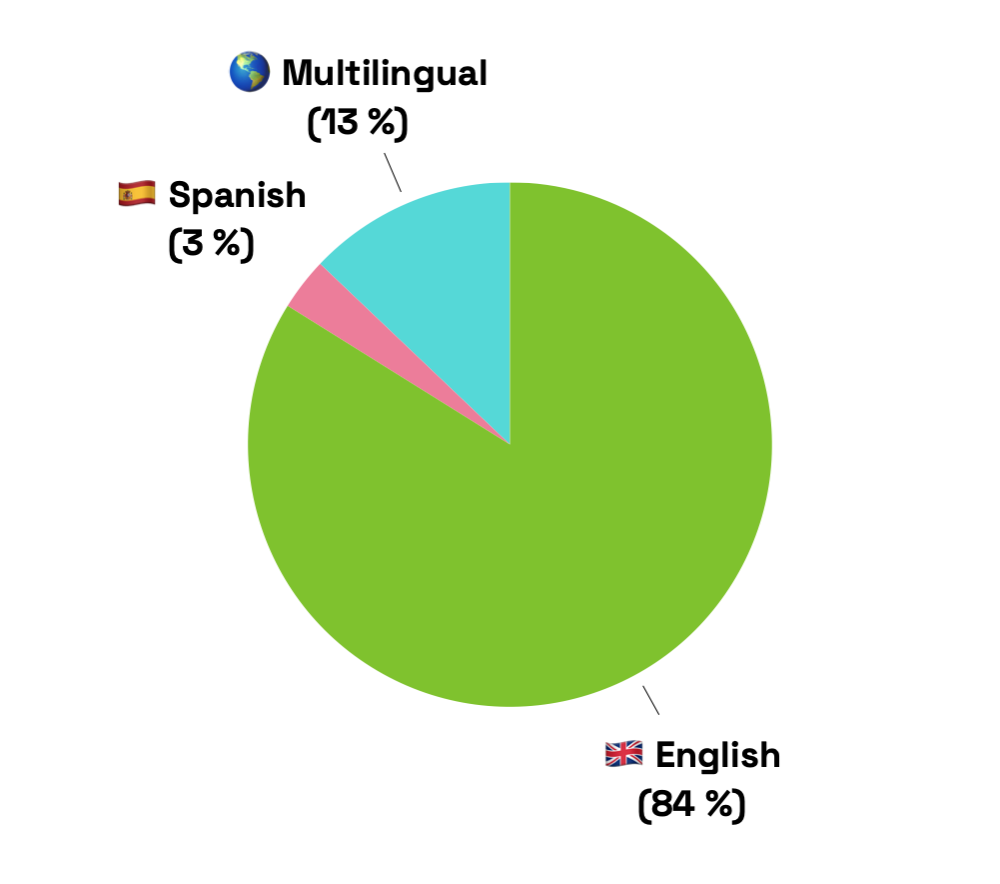}\label{fig:languages}}
  \hspace{0.6cm}
  \caption{Proportional statistics of the available disinformation classification datasets.}
\end{figure*}

\subsection{Disinformation, Misinformation}

With the advent and development of digital network media at the beginning of the 21st century, there has been a dynamic spread of unverified, inaccurate, or false information (raging from textual to audiovisual), which is referred to as information disorders. Information disorders as part of information pollution are thus in direct contrast to trustworthy content that is accurate, factually correct, verified, reliable, and up-to-date. According to the media and journalism theorist \cite{wardle2018need}, it is misleading to label information disorders with the umbrella term “fake news.” Although the definition of fake news is complicated, it is possible to define at least seven criteria that contribute to the contamination of information to such an extent that the use of the term information disorder is appropriate. 

Satire/parody as the least problematic form of information pollution and, therefore, a factor reducing the credibility of news content is on the one end of the seven-scale spectrum. In contrast, fictional content that was created for the intentional dissemination of false information lies at the other end. Wardle introduces a typology of the three main information disorders based on the seven criteria. The typology is established on the degree of truth/falsity and the intention to cause harm. Erroneous, inaccurate, or untrue content that is not intended to harm recipients because it reflects, for example, ignorance of the disseminator is referred to as misinformation. This term includes satire, parody, or misleading texts, images, or quotes. False or untrue content that is distributed to deceive or manipulate its recipients, whether for financial, ideological, political, social, or psychological reasons, is referred to as disinformation. This term includes malicious lies, fabricated information, disinformation campaigns, etc. Finally, true information disseminated with the intention to cause harm (for example, by revealing a person’s religion, sexual orientation, etc.) is referred to as malinformation.

The conceptual framework of individual information disorders in the professional literature is relatively inconsistent. Thus, part of the scientific community~\cite{fetzer2004disinformation} considers disinformation “misinformation with an attitude,” while attitude is the aforementioned deliberate deception of recipients. According to another approach~\cite{swire2020searching, wang2019systematic}, disinformation is part of misinformation because it is difficult to demonstrate the intention (not) to spread it. In both cases, the notion of misinformation encompasses the term disinformation. However, one can also encounter a more subtle division of individual forms of information disorders~\cite{meel2020fake}. In addition to the terms disinformation and misinformation, the authors also distinguish autonomous terms such as rumor, conspiracy, hoax, propaganda, opinion spam, false news (i.e., fake news), clickbait, satire, etc. Within the classification of information disorders, we can perceive disinformation and misinformation as overarching concepts because disinformation can take the form of clickbait, rumor, hoax, opinion spam, or conspiracy theory. Similarly, misinformation can be based on rumors or satire.

\subsection{Disinformation-related datasets}

\citet{d2021fake} have conducted a thorough study on fake news detection datasets. We highlight three of these based on the traction within the research community and direct the reader to this review for more detail.

\citet{wang2017liar} created the LIAR dataset with $12,836$ text excerpts of $6$ classes. Later, \citet{norregaard2019nela} published NELA-GT dataset containing 713,000 news articles belonging to $2$ classes. Lastly, \citet{slovikovskaya2020transfer} presented the FNC-1 dataset with $49,972$ news articles classified into $4$ labels. All these datasets are in English.

\citet{guo2022survey} have presented a survey of the current fact-checking datasets. Once again, we mention some of these below and refer the reader to the study for more detail.

First, \citet{mitra2015credbank} created the CredBank dataset with over $1,000$ English Tweets classified into $5$ labels. Multiple works followed, including the much larger Suspicious dataset~\cite{volkova2017separating} containing over $130,000$ English Tweets with $2$ assigned classes. Lastly, \citet{nakov2021clef} presented the CheckThat21-T1A dataset with over $17,000$ Tweets of $2$ classes. These Tweets come from multiple languages.

To capture the trends in the publishing intensity of these datasets, we include a plot of all the fake news detection datasets from \citet{d2021fake} by their year of origin in Figure~\ref{fig:dataset_growth}. This shows that the popularity of this task in the AI and NLP community is very much a recent phenomenon, corresponding to the general focus on the topic of disinformation in public. However, the sizeable collective excitement goes hand-in-hand with the inconsistency of the problem's framing and methodologies. This can be easily demonstrated with Figure~\ref{fig:classes}, which captures the distribution of these datasets by the pure number of labels into which they classify the articles. Furthermore, we see significant inconsistencies in the methodologies leading to these classifications if we go deeper. Some works~\cite{norregaard2019nela} derive the class based on the high-level credibility assessment of its source (i.e., whenever they deem a source problematic, they treat all its articles in this manner, leaving no room for exceptions). Others~\cite{wang2017liar, slovikovskaya2020transfer} treat the articles on an individual basis. Alongside, all of these differ in the specific features deducing the classification. Some consider the context of the article and editorial proprieties, while others only use the texts and its attributes.

Moreover, other major problematic characteristics of the dataset population emerge. Despite disinformation being a global threat, the vast majority of these datasets are in English only, as can be seen in Figure~\ref{fig:languages}. Alarmingly, most of the datasets did not include professionals or academics from the relevant fields, such as the media sciences and humanities in general. We believe that this calls for establishing a robust and uniform methodology for approaching the problem of disinformation holistically and an emphasis on developing datasets for non-English speaking regions with the oversight of relevant experts across domains and industries. We are addressing all of these in this in our work.

\subsection{Automated fake news detection}

The task of automated fake news detection has been usually approached by fine-tuning general-purpose language models, such as BERT~\cite{devlin-etal-2019-bert}, ELECTRA~\cite{clark2020electra}, or RoBERTa~\cite{abs-1907-11692}. Specific architectures for this task have been studied in the literature, too. \cite{reis2019supervised}, for instance, provide additional parameters such as political bias, the domain from the article's originating URL, and prior information about the domain as inputs to their model. \cite{singhal2019spotfake} create the first multi-modal architecture for this task as they combine the texts at the input with images included in the article.

\subsection{Automated fact-checking}

Architectures for automated fact-checking usually consist of an evidence retrieval module and a verification module~\cite{thorne2018fact}. Recent dense retrievers with learned representations and fast dot-product indexing~\cite{lewis2020retrieval, maillard2021multi} have shown strong performance, too. There have also been approaches considering multiple texts with potential evidence for the claims as a single evidence piece by concatenating them~\cite{luken2018qed, nie2019combining}. Later, an entailment model is employed to determine whether the article's text is supported or refuted by the evidence. We refer the reader to \cite{guo2022survey} for a concise overview of such methods.

\section{Trustworthiness Assessment Methodology}

Having familiarized ourselves with the current state of research, we concluded that the best way forward is to build upon the previous work and introduce a new language-agnostic methodology for classifying news articles. The primary motivation for this was the inability of prior approaches to fully reflect the complexity of the problem in terms of media studies and fully appreciate each article uniquely and independently of its source. We hope to provide better data for AI-based tools concerned with handling dubious news articles with this methodology. Below, we introduce the basic framework of our methodology. Its complete overview is available in Appendix~\ref{sec:appendix1}.

\subsection{Trustworthiness}

To strengthen clear division between the fake news detection and fact-checking tasks, our methodology focuses solely on the content aspects of the article. We hence do not reflect the truthfulness or context of the news, as we believe such practices fall under the latter task. These parameters on their own serve as robust evidence of an article being disinformative~\cite{alyt2021what}. 

Despite not determining the factualness, we can fully assess how deceptive it is and hence deem its trustworthiness. By omitting context and focusing solely on trustworthiness, we aim to improve the annotation process since there is no requirement for outside information and the class is final (i.e., unlike with methods employing truthfulness, no later information can reverse it).

\subsection{Classes}

Our methodology strays from the binary classification of true or fake news and allows for more granularity of class definitions because its sole focus is the assessment of trustworthiness.

To quantify trustworthiness, we propose $15$ negative linguistic attributes of an article (e.g., hate speech, clickbait title, logical fallacies) and $6$ positive ones (e.g., real author, references, objective profiling). With these, we define the following classes of trustworthiness:

\begin{enumerate}
    \item \textbf{Trustworthy}: These articles are credible and make no effort to deceive the reader. When relevant, they cite their sources of information and present the opinions of all involved parties. In terms of our framework, this means that they do not contain any negative attributes while having at least five positive ones.
    \item \textbf{Partially Trustworthy}: These articles still do not deceive their reader, but they often attempt to exaggerate the topic while making less effort to uphold journalistic norms. They often contain clickbait headlines and appeal to the readers' feelings. When translated into our framework, these include $2$ to $5$ negative attributes.
    \item \textbf{Misleading}: These articles contain deception outside the boundaries of pure conspiracies. These articles often alter the framings of the news to fit their agenda. Any article containing 6 to 8 negative attributes belongs to this class.
    \item \textbf{Manipulative}: These articles strive to manipulate their reader. Hence, their arguments often use conspiracy narratives. They contain over $8$ negative attributes or $3$, especially problematic ones, such as using conspiracies or hate speech.
\end{enumerate}

\section{Verifee Dataset}
\label{sec:dataset}

Using our newly introduced methodology, we collected a dataset of over $10,000$ Czech news articles classified into the just-described categories. Apart from the class, each entry in the dataset consists of the article's text, HTML source, title, description, authors, keywords, source name, URL, covered controversial topics, and included images. We open-source the dataset at \url{https://verifee.ai/research} under a custom license\footnote{Our license — building on top of Creative Commons BY-NC-SA (\url{https://creativecommons.org/licenses/by-nc-sa/2.0/}) — is available at \url{https://www.verifee.ai/files/license.pdf}.}. We provide pre-defined train ($80~\%$), validation ($10~\%$), and ($10~\%$) testing splits that have been assigned randomly. Below, we describe the process of the dataset's collection.

\subsection{Scraping and Pre-processing}
Initially, we assembled a collection of almost $94,000$ articles by scraping URLs of 45 Czech news sources obtained from Common Crawl\footnote{\url{https://commoncrawl.org}}. These sources included mainstream journalistic websites, tabloids, independent news outlets, and websites that are part of the disinformation ecosystem~\cite{stetka2021nobody}, capturing the full scope of journalistic content in the Czech Republic. Their complete list can be found in Appendix~\ref{sec:appendix2}.

\subsubsection{Enrichment}

Next, we determined the category (opinion, interview, general) and the topic (general, sport, economics, hobby, tabloid) of each article through pattern matching. We detected mentions of any controversial topics relevant to the Czech media context (Russia, Covid-19, EU, NATO, USA, and migration) similarly. Additionally, we ascertained whether the article disposes of a real author via an out-of-the-box Named Entity Recognition model~\cite{sido2021czert} for the Czech language.

\subsubsection{Filtering}

We applied multiple filters and balancing mechanisms to mitigate deficiencies caused by inherent flaws in Common Crawl, which reduced the dataset's size from $94,000$ to $10,000$ items. This way, we also ensured that the data is as representative of the Czech news ecosystem and as diverse as possible. The factors used for filtering were:

\begin{itemize}
    \item \textbf{Length of the text}: Only articles with a length between $400$ and $10,000$ characters were included.
    \item \textbf{Category}: For mainstream media, we filtered opinion pieces. However, we kept these for alternative news sources, as the line between reporting and conveying opinion is often blurred here. Interviews were excluded in both cases.
    \item \textbf{Source}: We selected articles in such a way that all sources are as balanced as possible, no matter their actual distribution in the media ecosystem.
    \item \textbf{Topic}: Articles concerning hobbies and sports each form only 5\% of the dataset, as they typically are not connected to disinformation. The rest of the topics (general, economic, and tabloid) each form $30~\%$ of the dataset.
    \item \textbf{Controversial topics}: We made sure to balance the coverage of controversial topics by including the same number of such articles from mainstream and alternative or extremely opinionated news sources.
\end{itemize}

\subsection{Annotations Organization}

We conducted two rounds of annotation. The first round involved $7,347$ unique articles, where just the class was denoted to each article. The second round included $2,655$ unique articles. This time, annotators were asked to provide both the class and flag any problematic attributes of each article defined in our methodology. This enabled us to examine the importance of the various metrics in the methodology. Every annotator was assigned $40$ articles in both rounds.

\subsubsection{Annotators}

All the raters were students of journalism who are native speakers of the Czech language. They thus had a more advanced understanding of the topic of news credibility than the general population. We have to note that due to their age~\cite{Peltzman2019PoliticalIO} and education~\cite{SCOTT2022102471}, their possible bias toward more progressive/liberal schools of thought may have influenced the rating of topics in corresponding areas. We briefed all the annotators on an extensive seminar, provided all of them with detailed materials, and encouraged them to come forward with any problems.

\subsubsection{Platform}

We adapted the open-source tool Doccano\footnote{\url{https://doccano.github.io/doccano/}} for our task and used it in the collection. Inside the application, annotators were presented with one article at a time in its HTML form with all images included. The platform allowed the user to add necessary tags and comments to each piece.

\subsubsection{Source Identity Masking}

We masked any elements in the article that would enable the annotators to identify the source or author of the text. Specifically, we replaced their mentions with placeholders. As a result, annotators' media and author preferences could not influence their evaluation.

\subsubsection{Monitoring}
\label{subsubsec:monitoring}

We tracked the raters' activity on the platform during the annotation process. This includes the time they spent annotating each item and the exact timeline of their activity. In the second wave, $10~\%$ of each annotator's assigned set of articles was the same for later evaluation of the inter-annotator agreement. We selected these before the collection process and gave them our ground-truth annotations to ensure that the class distribution within this subset is balanced. We present the results of speed and inter-annotator agreement analyses later in this section.

\subsection{Data Analysis}

In this section, we present statistics of the newly-collected dataset. Overall, it contains $10229$ articles spanning $60$ Czech sources. A detailed list of all sources, their respective item counts, and class distributions can be found in Appendix~\ref{sec:appendix2}. Plotted in Figure~\ref{fig:annotators_speed} is the distribution of their times spent on individual articles. By average, annotators spent $2,95$ minutes ($177$ seconds) on a single article, which indicates reasonable time allocation.

\begin{table}[t!]
\begin{tabular}{l|l}
\textbf{Class} & \textbf{Number of articles}\\ \hline
Trustworthy            & 3534                  \\
Partially trustworthy        & 2577                  \\
Misleading & 1526     \\    
Manipulative & 1859     \\    
Unclassifiable & 733     \\    
\end{tabular}
\label{tab:class_distrib}
\caption{Distribution of article items per annotated credibility class}
\end{table}

We pay close attention to the per-source class distributions and ensure that the general tendencies in annotations match the Czech media space analyses studying the high-level credibility of news outlets. State-owned media (ČTK, ČT24, and iROZHLAS) and local newspapers (Jihlavské listy and Mostecké listy) have a majority of their stories classified as 'Trustworthy.' Articles from private media outlets (Seznam Zprávy, iDnes, Deník) are also most often classified as 'Trustworthy.' This time, however, other classes are more prominent. Openly left-wing (A2larm) or right-wing (Echo 24 and Forum24) sources have more items classified as misleading or manipulative in comparison to their counterparts without distinctive political tendencies. The 'Partially trustworthy' news stories occur the most by tabloid news sites (Blesk, Aha!, Extra.cz).

We can see the disinformative news sites (Aeronet, Protiproud, Skrytá pravda) on the other side of the spectrum, as their articles get exceedingly labeled as 'misleading' and 'manipulative.' % Nevertheless, items from two outlets considered some of the primary sources of disinformation in the Czech Republic (Sputnik News and ParlamentníListy.cz) did not follow this trend. Their content is mainly classified as 'Trustworthy' or 'Partially trustworthy,' though, the number of 'Misleading' and 'Manipulative' news stories from these sources is still dramatically larger than that of mainstream media outlets. Although more investigation into this matter needs to be done, it seems that these two sources have a specific publishing strategy. They produce a large mass of content, out of which only a small part is manipulative. The vast amount of entirely credible journalistic work probably serves to establish trust with the readers to make them more vulnerable to the manipulative pieces.

Overall, we can see that the high-level patterns in the annotations match the news sources' characteristics, as described in media science literature~\cite{stetka2021nobody}.

\begin{figure}[t!]
\begin{center}
\includegraphics[width=1\linewidth]{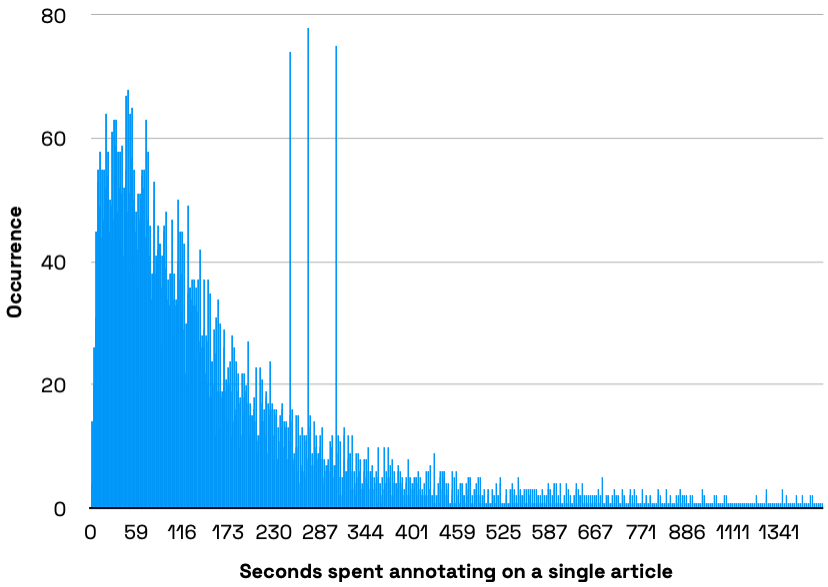}
\end{center}
\caption{Distribution of the times spent on single articles by annotators. The x-axis denotes the number of seconds and the y-axis the count of respective occurrences.}
\label{fig:annotators_speed}
\end{figure}

\subsubsection{Inter-annotator Agreement}
\label{subsub:agreement}

We gathered data for determining the inter-annotator agreement by assigning $4$ duplicated articles ($10~\%$) to each annotator in the second annotation wave. We computed Randolph's Kappa \cite{randolph2010kappa} and arrived on a value of $0.615$, which corresponds to a moderate agreement~\cite{mchugh2012interrater}. This indicates a general understanding of the task and rather. At the same time, there is room for the latter filtering of problematic annotators, who can be spotted by largely deviating in categorizing these duplicated articles.

\section{Experimental Results}

To establish initial benchmarks, we trained models of $3$ distinct architectures for the task of news trustworthiness classification on our dataset. Herein, we present the experimental setup and the results.

% \usepackage{multirow}
\begin{table*}[h]
\begin{center}
\begin{tabular}{l|llll|l}
\multirow{2}{*}{\textbf{Model}} & \multicolumn{4}{c|}{\textbf{F-1 score}} & \textbf{Overall average}                                                                 \\
                                & Trustworthy  & Partially credible & Misleading   & Manipulative & \textbf{macro F-1 score} \\ \hline
TF-IDF SVM                      & \multicolumn{1}{l|}{ 0.52} & \multicolumn{1}{l|}{ 0.40}          & \multicolumn{1}{l|}{0.35} &    \multicolumn{1}{l|}{\textbf{0.68}}          & \multicolumn{1}{l}{0.49}                    \\
FastText LG                    & \multicolumn{1}{l|}{\textbf{0.58}} & \multicolumn{1}{l|}{0.28}          & \multicolumn{1}{l|}{0.14} &   \multicolumn{1}{l|}{0.60}          & \multicolumn{1}{l}{0.40}                     \\
Czert (BERT)                    & \multicolumn{1}{l|}{0.55} & \multicolumn{1}{l|}{\textbf{0.47}}          & \multicolumn{1}{l|}{\textbf{0.44}} &   \multicolumn{1}{l|}{0.61}          & \multicolumn{1}{l}{\textbf{0.52}}                       
\end{tabular}
\end{center}
\caption{Verifee dataset benchmarks using three classification architectures: TF-IDF, FastText, and Czert (a BERT model for the Czech language). We report the F-1 score on the testing split.}
\label{tab:results}
\end{table*}

\subsection{Experimental setting}

We experimented with three model architectures: Term frequency-inverse document frequency (TF-IDF)-based Support Vector Machines (SVM) classifier~\cite{tfidf, hearst1998support}, FastText classifier~\cite{joulin2017bag}, and BERT~\cite{devlin2019bert}. We first describe the data preparation procedure and later review the configuration of each employed model.

\subsection{Data Preparation}

We follow the pre-defined configuration of train, test, and validation sets described in Section~\ref{sec:dataset}. To balance the dataset, we removed all duplicates used to measure kappa and randomly selected a sample of 1400 articles from each credibility class. We do not employ the $733$ items categorized as 'Unclassifiable'. We insert the article's title and body concatenated with a period as the input to each evaluated model.

\subsubsection{TF-IDF}

We first trained a TF-IDF-based Support Vector Machines model. Therein, the text of an article is first vectorized using TF-IDF. Then, the SVM  model classifies the content as one of the $4$ credibility classes. We used the scikit-learn library~\cite{scikit} for its implementation. 

We kept the model's vocabulary unfiltered by setting its \texttt{min\_df} and \texttt{max\_df} parameters to $1$. For the SVM, we used Radial basis function kernel and Regularization parameter set to 1.

\subsubsection{FastText}

Another architecture we included in the benchmarking is a FastText classifier. In this pipeline, each article text is first tokenized using nltk. An article vector is then obtained by averaging FastText word embeddings of these tokens. We use one of FastText library's pre-trained models~\cite{grave2018learning}, which was trained on the Czech corpus from Common Crawl and Wikipedia. Lastly, a one-vs-rest logistic regression~\cite{logisticregression} classifies the article.

As for the hyperparameters, we used L2 penalty term combined with regulization set to $1$.

\subsubsection{BERT}

Lastly, we also tried the Czert model~\cite{sido2021czert}, which utilizes the BERT architecture but is trained on the Czech national corpus~\cite{krenczech}, Czech Wikipedia, and a scrape of Czech news.

We fine-tuned the model for the purposes of our classification task using the Cross-entropy loss and the learning rate set at $3 * 10^{-5}$. We fine-tuned the model for $4$ epochs and batch size set to $32$.

\subsection{Results}

We present the per-class and overall F-1 score results in Table~\ref{tab:results}. As can be observed, the scores distinctively differ across classes. Upon closer inspection, both TF-IDF SVM and FastText LG models perform better on the classes at either pole of the trustworthiness spectrum (i.e., 'Trustworthy' and 'Manipulative'), but underperform at the middle ones, resulting in overall testing F-1 scores of $0.49$ and $0.40$, respectively. We expect that the poor performance of the FastText LG model is caused due to the inability to construct granular representation that is necessary for our task, which cannot be resembled by averaged word embeddings. Despite sligthly losing on the pole classes, the Czert (BERT) model resembles best robustness across the spectrum. It achieves an overall testing F-1 score of $0.52$

\section{Ethical Discussion and Limitations}

Due to the high-impact nature of the solved task, we consider it appropriate to review ethical considerations that have been made during this research project. Additionally, we outline further steps we are making to ensure safety and transparency even beyond publication and recommendations for build-up work.

First, let us focus on the presence of biases in the data. Despite avoiding this statistical phenomenon being practically impossible, we put extensive procedures in place even at the very start of the project. By inviting media researchers into our core team, we wanted to minimize misunderstandings and mistakes that scientists from the field of computational linguistics could easily make when assembling the methodology for the task of trustworthiness assessment due to their limited knowledge of the current literature and theory in the area of journalism. Prior to the data annotation, we invited scholars in media studies and journalists from the industry to a series of workshops, where we asked them to submit feedback and discuss the methodology. Based on the assembled comments, we kept updating it until a general consensus was reached. In terms of the annotation process itself, multiple safeguards have been employed to prevent the annotators' source or author preference.

Second, let us shift towards the ethics of using any technology built around this data in the wild. We want to stress that anyone using this dataset for the purposes of creating a trustworthiness classification system should provide transparent information to the users that this process is automatic and hence faulty to a certain extent. We must note that it is yet unclear how models trained on this data generalize for future articles (i.e., news about topics and events they have not encountered in the training set) and news sources that were not included in the training set. A study into these should be conducted prior to making this technology available unrestrictedly to the public.

Despite bearing these safety questions in mind is crucial, we believe that such systems can eventually be great assistive tools for people reading news stories online. The potential benefits of such technology should support initiatives into safeguarding it first and hence establishing public and academic trust.

\section{Conclusion}

This work presents a novel methodology for classifying news article trustworthiness and presents a dataset of nearly $10,000$ Czech news articles with respective annotations. Unlike previous methods that classify all texts from a given media outlet with the same class, we treat the articles on an individual level. The high inter-annotator agreement shows that our methodology constitutes a good feature-based framework, leaving little to no room for personal annotators' inducement.

To the best of our knowledge, we are the first to include media and computer science researchers in the core team when developing a similar dataset. Additionally, all of our annotators were journalism students. As our methodology underwent extensive feedback loops with professionals in the industry, we hope to establish a new interdisciplinary standard for future related works to follow. 

We provide benchmark results on our dataset using $3$ different classifier architectures and obtain promising results. We open-source both the methodology and data and encourage researchers to undertake similar initiatives in new languages and social contexts, especially low-resourced ones. Since the framework derives all parameters based on the text contents, it is language-agnostic. Hence, minimal additional methodological work is necessary before new annotations.

In future work, we intend to study the generalization abilities of systems trained using this data and the application of task-specific architectures. Moreover, we wish to further explore the potential of multimodality that our dataset offers and analyze the attached images.

\section*{Acknowledgements}

This paper was supported by the Technology Agency of the Czech Republic under grant No. TL05000057 "The Signal and the Noise in the Era of Journalism 5.0 - A Comparative Perspective of Journalistic Genres of Automated Content".

%%
%% Define the bibliography file to be used
\bibliography{main}

\clearpage

\appendix

\section{Annotation Methodology and Annotators Instructions}
\label{sec:appendix1}

\subsection{Annotation instructions}
Each class is defined by the positive aspects it contains and the negative aspects it can and cannot contain. When annotating, we start with the most trustworthy class (credible). We then move down a class whenever an article does not meet the requirements of the current class, for example when it contains too many permissible negative aspects or contains a negative aspect that must not occur in that class.

\subsection{Trustworthiness classes}
\subsubsection{Trustworthy}
\textbf{Positive aspects contained in the article \mbox{(min. 5):}}
\begin{itemize}
    \item Citation of relevant authorities on the topic, representing credible institutions
    \item Views of all interested parties
    \item Facts presented within the context
    \item Grammatical correctness, without overtly expressive language 
    \item An identifiable author
    \item Undistorted data
\end{itemize}
\noindent
\textbf{Negative aspects contained in the article \mbox{(max. 1):}}
\begin{itemize}
    \item Missing citations
    \item Unrepresented views of opposing parties
    \item Facts presented without a context
    \item Grammatically incorrect or overtly expressive language 
    \item Unidentifiable author
    \item Distorted data
\end{itemize}
\noindent
\textbf{Negative aspects that must not appear in the article:}
\begin{itemize}
    \item Clickbait
    \item Hate speech
    \item An attack on an opinion opponent without justification
    \item Manipulating the reader
    \item Conspiracy theories
    \item Appeal to emotion
    \item Logical fallacies
    \item Tabloid language
\end{itemize}

\subsubsection{Partially Trustworthy}
\textbf{Positive aspects contained in the article:}
\begin{itemize}
    \item Grammatical correctness, without overtly expressive language 
    \item Undistorted data
\end{itemize}
\noindent
\textbf{Negative aspects contained in the article \mbox{(2-5):}}
\begin{itemize}
    \item Missing citations
    \item Unrepresented views of opposing parties
    \item Facts presented without a context
    \item Grammatically incorrect or overtly expressive language 
    \item Unidentifiable author
    \item Distorted data
    \item Clickbait
    \item Appeal to emotion
    \item Tabloid language
\end{itemize}
\noindent
\textbf{Negative aspects that must not appear in the article:}
\begin{itemize}
    \item Hate speech
    \item An attack on an opinion opponent without justification
    \item Manipulating the reader
    \item Conspiracy theories
    \item Logical fallacies
\end{itemize}

\subsubsection{Misleading}
\textbf{Positive aspects contained in the article:}
\begin{itemize}
    \item[] \textit{None need to be present}
\end{itemize}
\noindent
\textbf{Negative aspects contained in the article \mbox{(6-7):}}
\begin{itemize}
    \item Missing citations
    \item Unrepresented views of opposing parties
    \item Facts presented without a context
    \item Grammatically incorrect or overtly expressive language 
    \item Unidentifiable author
    \item Distorted data
    \item Clickbait
    \item Appeal to emotion
    \item Tabloid language
    \item Logical fallacies
    \item An attack on an opinion opponent without justification
\end{itemize}
\noindent
\textbf{Negative aspects that must not appear in the article:}
\begin{itemize}
    \item Hate speech
    \item Manipulating the reader
    \item Conspiracy theories
\end{itemize}

\subsubsection{Manipulative}
\textbf{Positive aspects contained in the article:}
\begin{itemize}
    \item[] \textit{None need to be present}
\end{itemize}
\noindent
\textbf{Negative aspects contained in the article:}
\begin{itemize}
    \item[] \textit{It either contains 8+ negative aspects:}
    \item Missing citations
    \item Unrepresented views of opposing parties
    \item Facts presented without a context
    \item Grammatically incorrect or overtly expressive language 
    \item Unidentifiable author
    \item Distorted data
    \item Clickbait
    \item Appeal to emotion
    \item Tabloid language
    \item Logical fallacies
    \item An attack on an opinion opponent without justification
    \item[] \textit{Or it contains any of these 3 aspects:}
    \item Hate speech
    \item Manipulating the reader
    \item Conspiracy theories
\end{itemize}
\noindent
\textbf{Negative aspects that must not appear in the article:}
\begin{itemize}
    \item[] \textit{All negative aspects can be present}
\end{itemize}

\subsection{Resolving unclassifiable articles and errors}
\subsubsection{Unclassifiable articles}
Articles that, due to their length or structure, cannot be classified according to this methodology (or do not have sufficient content to allow the aspects mentioned
to be analysed) must be labeled as unclassifiable. This may include one-sentence flash news announcements, paywall texts and
others. This allows them to be filtered out and not corrupt the rest of the annotated data.

\subsubsection{Errors}
In the case that an error with the platform or an uncertainty with an article is encounter, we fully encourage annotators to report those issues through comment functionality on the Doccano platform. Our team will do their best to resolve any problem and clarify any ambiguity.

\section{Detailed news source statistics}
\label{sec:appendix2}

\textit{Continued on the next page.}

% \usepackage{multirow}
\begin{table*}[]
\centering
\small
\begin{tabular}{l|p{2cm}p{2cm}p{2cm}p{2cm}p{2cm}}
\multirow{2}{*}{\textbf{News source}} & \multicolumn{5}{c}{\textbf{Article items per class}}                                                                                                                                                              \\ \cline{2-6} 
                                      & \multicolumn{1}{l|}{\textbf{Trustworthy}} & \multicolumn{1}{l|}{\textbf{Part. trustworthy}} & \multicolumn{1}{l|}{\textbf{Misleading}} & \multicolumn{1}{l|}{\textbf{Manipulative}} & \textbf{Unclassifiable} \\ \hline
A2larm                                & \multicolumn{1}{l|}{158}                  & \multicolumn{1}{l|}{106}                            & \multicolumn{1}{l|}{49}                  & \multicolumn{1}{l|}{13}                    & 20                      \\ \hline
AC24                                  & \multicolumn{1}{l|}{22}                   & \multicolumn{1}{l|}{40}                             & \multicolumn{1}{l|}{45}                  & \multicolumn{1}{l|}{26}                    & 22                      \\ \hline
Aeronet                               & \multicolumn{1}{l|}{6}                    & \multicolumn{1}{l|}{19}                             & \multicolumn{1}{l|}{69}                  & \multicolumn{1}{l|}{347}                   & 9                       \\ \hline
Aha!                                  & \multicolumn{1}{l|}{20}                   & \multicolumn{1}{l|}{73}                             & \multicolumn{1}{l|}{41}                  & \multicolumn{1}{l|}{7}                     & 16                      \\ \hline
Aktuálně                              & \multicolumn{1}{l|}{234}                  & \multicolumn{1}{l|}{107}                            & \multicolumn{1}{l|}{33}                  & \multicolumn{1}{l|}{5}                     & 38                      \\ \hline
Blesk                                 & \multicolumn{1}{l|}{40}                   & \multicolumn{1}{l|}{132}                            & \multicolumn{1}{l|}{35}                  & \multicolumn{1}{l|}{5}                     & 6                       \\ \hline
Brněnský deník                        & \multicolumn{1}{l|}{27}                   & \multicolumn{1}{l|}{10}                             & \multicolumn{1}{l|}{2}                   & \multicolumn{1}{l|}{0}                     & 4                       \\ \hline
CNN Prima News                        & \multicolumn{1}{l|}{232}                  & \multicolumn{1}{l|}{86}                             & \multicolumn{1}{l|}{13}                  & \multicolumn{1}{l|}{2}                     & 16                      \\ \hline
CZ24 News                             & \multicolumn{1}{l|}{16}                   & \multicolumn{1}{l|}{28}                             & \multicolumn{1}{l|}{12}                  & \multicolumn{1}{l|}{21}                    & 2                       \\ \hline
Czech free press                      & \multicolumn{1}{l|}{3}                    & \multicolumn{1}{l|}{15}                             & \multicolumn{1}{l|}{12}                  & \multicolumn{1}{l|}{10}                    & 2                       \\ \hline
Deník                                 & \multicolumn{1}{l|}{58}                   & \multicolumn{1}{l|}{14}                             & \multicolumn{1}{l|}{4}                   & \multicolumn{1}{l|}{0}                     & 2                       \\ \hline
Deník N                               & \multicolumn{1}{l|}{28}                   & \multicolumn{1}{l|}{8}                              & \multicolumn{1}{l|}{4}                   & \multicolumn{1}{l|}{1}                     & 9                       \\ \hline
Deník Referendum                      & \multicolumn{1}{l|}{192}                  & \multicolumn{1}{l|}{51}                             & \multicolumn{1}{l|}{22}                  & \multicolumn{1}{l|}{6}                     & 5                       \\ \hline
E-republika                           & \multicolumn{1}{l|}{4}                    & \multicolumn{1}{l|}{7}                              & \multicolumn{1}{l|}{13}                  & \multicolumn{1}{l|}{16}                    & 2                       \\ \hline
Echo 24                               & \multicolumn{1}{l|}{203}                  & \multicolumn{1}{l|}{84}                             & \multicolumn{1}{l|}{21}                  & \multicolumn{1}{l|}{2}                     & 6                       \\ \hline
Euro                                  & \multicolumn{1}{l|}{14}                   & \multicolumn{1}{l|}{8}                              & \multicolumn{1}{l|}{1}                   & \multicolumn{1}{l|}{0}                     & 0                       \\ \hline
Euro Zprávy                           & \multicolumn{1}{l|}{55}                   & \multicolumn{1}{l|}{18}                             & \multicolumn{1}{l|}{4}                   & \multicolumn{1}{l|}{0}                     & 5                       \\ \hline
Extra.cz                              & \multicolumn{1}{l|}{86}                   & \multicolumn{1}{l|}{229}                            & \multicolumn{1}{l|}{116}                 & \multicolumn{1}{l|}{37}                    & 25                      \\ \hline
Forum24                               & \multicolumn{1}{l|}{147}                  & \multicolumn{1}{l|}{64}                             & \multicolumn{1}{l|}{39}                  & \multicolumn{1}{l|}{21}                    & 13                      \\ \hline
Globe 24                              & \multicolumn{1}{l|}{15}                   & \multicolumn{1}{l|}{7}                              & \multicolumn{1}{l|}{2}                   & \multicolumn{1}{l|}{0}                     & 0                       \\ \hline
Haló noviny                           & \multicolumn{1}{l|}{14}                   & \multicolumn{1}{l|}{14}                             & \multicolumn{1}{l|}{16}                  & \multicolumn{1}{l|}{4}                     & 3                       \\ \hline
Hospodářské noviny                    & \multicolumn{1}{l|}{34}                   & \multicolumn{1}{l|}{14}                             & \multicolumn{1}{l|}{4}                   & \multicolumn{1}{l|}{6}                     & 78                      \\ \hline
INFO.cz                               & \multicolumn{1}{l|}{18}                   & \multicolumn{1}{l|}{19}                             & \multicolumn{1}{l|}{4}                   & \multicolumn{1}{l|}{2}                     & 14                      \\ \hline
Jihlavské listy                       & \multicolumn{1}{l|}{26}                   & \multicolumn{1}{l|}{3}                              & \multicolumn{1}{l|}{1}                   & \multicolumn{1}{l|}{0}                     & 3                       \\ \hline
Lidovky.cz                            & \multicolumn{1}{l|}{5}                    & \multicolumn{1}{l|}{19}                             & \multicolumn{1}{l|}{4}                   & \multicolumn{1}{l|}{3}                     & 30                      \\ \hline
MediaGuru                             & \multicolumn{1}{l|}{24}                   & \multicolumn{1}{l|}{18}                             & \multicolumn{1}{l|}{3}                   & \multicolumn{1}{l|}{0}                     & 1                       \\ \hline
Metro                                 & \multicolumn{1}{l|}{135}                  & \multicolumn{1}{l|}{53}                             & \multicolumn{1}{l|}{5}                   & \multicolumn{1}{l|}{0}                     & 9                       \\ \hline
Mostecké listy                        & \multicolumn{1}{l|}{22}                   & \multicolumn{1}{l|}{3}                              & \multicolumn{1}{l|}{1}                   & \multicolumn{1}{l|}{0}                     & 1                       \\ \hline
NWOO                                  & \multicolumn{1}{l|}{8}                    & \multicolumn{1}{l|}{35}                             & \multicolumn{1}{l|}{37}                  & \multicolumn{1}{l|}{63}                    & 15                      \\ \hline
Novinky.cz                            & \multicolumn{1}{l|}{67}                   & \multicolumn{1}{l|}{67}                             & \multicolumn{1}{l|}{13}                  & \multicolumn{1}{l|}{3}                     & 15                      \\ \hline
Nová republika                        & \multicolumn{1}{l|}{5}                    & \multicolumn{1}{l|}{34}                             & \multicolumn{1}{l|}{51}                  & \multicolumn{1}{l|}{56}                    & 8                       \\ \hline
Outsider Media                        & \multicolumn{1}{l|}{94}                   & \multicolumn{1}{l|}{119}                            & \multicolumn{1}{l|}{162}                 & \multicolumn{1}{l|}{235}                   & 92                      \\ \hline
Parlamentní­ Listy                    & \multicolumn{1}{l|}{279}                  & \multicolumn{1}{l|}{269}                            & \multicolumn{1}{l|}{138}                 & \multicolumn{1}{l|}{91}                    & 33                      \\ \hline
Peak.cz                               & \multicolumn{1}{l|}{122}                  & \multicolumn{1}{l|}{49}                             & \multicolumn{1}{l|}{12}                  & \multicolumn{1}{l|}{1}                     & 6                       \\ \hline
Proti Proud                           & \multicolumn{1}{l|}{15}                   & \multicolumn{1}{l|}{41}                             & \multicolumn{1}{l|}{102}                 & \multicolumn{1}{l|}{301}                   & 23                      \\ \hline
Raptor TV                             & \multicolumn{1}{l|}{1}                    & \multicolumn{1}{l|}{4}                              & \multicolumn{1}{l|}{2}                   & \multicolumn{1}{l|}{3}                     & 1                       \\ \hline
Reflex                                & \multicolumn{1}{l|}{1}                    & \multicolumn{1}{l|}{1}                              & \multicolumn{1}{l|}{3}                   & \multicolumn{1}{l|}{1}                     & 11                      \\ \hline
Rukojmí                               & \multicolumn{1}{l|}{21}                   & \multicolumn{1}{l|}{52}                             & \multicolumn{1}{l|}{117}                 & \multicolumn{1}{l|}{290}                   & 12                      \\ \hline
Seznam Zprávy                         & \multicolumn{1}{l|}{181}                  & \multicolumn{1}{l|}{51}                             & \multicolumn{1}{l|}{13}                  & \multicolumn{1}{l|}{1}                     & 8                       \\ \hline
Skrytá Pravda                         & \multicolumn{1}{l|}{6}                    & \multicolumn{1}{l|}{17}                             & \multicolumn{1}{l|}{64}                  & \multicolumn{1}{l|}{169}                   & 11                      \\ \hline
Sputnik ČR                            & \multicolumn{1}{l|}{219}                  & \multicolumn{1}{l|}{296}                            & \multicolumn{1}{l|}{124}                 & \multicolumn{1}{l|}{43}                    & 34                      \\ \hline
Stars 24                              & \multicolumn{1}{l|}{27}                   & \multicolumn{1}{l|}{42}                             & \multicolumn{1}{l|}{15}                  & \multicolumn{1}{l|}{3}                     & 2                       \\ \hline
Svobodné noviny                       & \multicolumn{1}{l|}{13}                   & \multicolumn{1}{l|}{23}                             & \multicolumn{1}{l|}{45}                  & \multicolumn{1}{l|}{77}                    & 8                       \\ \hline
TN.cz                                 & \multicolumn{1}{l|}{213}                  & \multicolumn{1}{l|}{204}                            & \multicolumn{1}{l|}{38}                  & \multicolumn{1}{l|}{3}                     & 17                      \\ \hline
Týden                                 & \multicolumn{1}{l|}{54}                   & \multicolumn{1}{l|}{14}                             & \multicolumn{1}{l|}{4}                   & \multicolumn{1}{l|}{0}                     & 4                       \\ \hline
VOX Populi                            & \multicolumn{1}{l|}{4}                    & \multicolumn{1}{l|}{14}                             & \multicolumn{1}{l|}{60}                  & \multicolumn{1}{l|}{72}                    & 22                      \\ \hline
Zvědavec                              & \multicolumn{1}{l|}{6}                    & \multicolumn{1}{l|}{6}                              & \multicolumn{1}{l|}{6}                   & \multicolumn{1}{l|}{7}                     & 5                       \\ \hline
iDnes.cz                              & \multicolumn{1}{l|}{93}                   & \multicolumn{1}{l|}{39}                             & \multicolumn{1}{l|}{11}                  & \multicolumn{1}{l|}{1}                     & 19                      \\ \hline
iROZHLAS                              & \multicolumn{1}{l|}{254}                  & \multicolumn{1}{l|}{69}                             & \multicolumn{1}{l|}{13}                  & \multicolumn{1}{l|}{1}                     & 18                      \\ \hline
ČT24                                  & \multicolumn{1}{l|}{246}                  & \multicolumn{1}{l|}{36}                             & \multicolumn{1}{l|}{5}                   & \multicolumn{1}{l|}{3}                     & 35                      \\ \hline
ČTK                                   & \multicolumn{1}{l|}{38}                   & \multicolumn{1}{l|}{2}                              & \multicolumn{1}{l|}{2}                   & \multicolumn{1}{l|}{0}                     & 1                       \\ \hline
Časopis Šifra                         & \multicolumn{1}{l|}{11}                   & \multicolumn{1}{l|}{10}                             & \multicolumn{1}{l|}{10}                  & \multicolumn{1}{l|}{8}                     & 3                       \\ \hline
Česko Aktuálně                        & \multicolumn{1}{l|}{36}                   & \multicolumn{1}{l|}{42}                             & \multicolumn{1}{l|}{34}                  & \multicolumn{1}{l|}{35}                    & 8                      
\end{tabular}
\caption{Distribution of classes per individual news sources}
\end{table*}

\end{document}